\documentclass[runningheads]{llncs}
\usepackage[T1]{fontenc}
\usepackage{graphicx}
\usepackage{booktabs, multirow} 
\usepackage{soul}
\begin{document}
\title{Beating Transformers using Synthetic Cognition}
%
%
\author{
Alfredo Ibias\inst{1}\orcidID{0000-0002-3122-4272} \and
Miguel Rodriguez-Galindo\inst{1}\orcidID{0000-0003-2893-916X} \and
Hector Antona\inst{1} \and
Guillem Ramirez-Miranda\inst{1}\orcidID{0000-0003-2741-3705} \and
Enric Guinovart\inst{1} 
}
\authorrunning{
A. Ibias et al.
}
%
\institute{
Avatar Cognition, Barcelona, Spain\\
\email{\{alfredo, miguel, hector, guillem, enric\}@avatarcognition.com}
}
\maketitle              
\begin{abstract}
The road to Artificial General Intelligence goes through the generation of context-aware reactive behaviors, where the Transformer architecture has been proven to be the state-of-the-art. However, they still fail to develop reasoning. Recently, a novel approach for developing cognitive architectures, called Synthetic Cognition, has been proposed and implemented to develop instantaneous reactive behavior. In this study, we aim to explore the use of Synthetic Cognition to develop context-aware reactive behaviors. We propose a mechanism to deal with sequences for the recent implementation of Synthetic Cognition, and test it against DNA foundation models in DNA sequence classification tasks. In our experiments, our proposal clearly outperforms the DNA foundation models, obtaining the best score on more benchmark tasks than the alternatives. Thus, we achieve two goals: expanding Synthetic Cognition to deal with sequences, and beating the Transformer architecture for sequence classification.

\end{abstract}
\section{Introduction}
On the road to Artificial General Intelligence (AGI) there are some fundamental steps. The first one, widely achieved by most Artificial Intelligence (AI) methods, is the development of instantaneously reactive behaviour. This is what we call \emph{pattern matching}, as any instantaneously reactive behaviour consists of matching the pattern of external inputs (also called state) to one of its stored ones, in order to produce the associated response (also called action). However, these behaviours, being purely instantaneous, do not account for the time context, which comes in the second step: the development of episodic reactive behaviours. These behaviours are also based on pattern matching, but this time, taking into account the previous inputs. These behaviours are, in the end, context-aware reactive behaviours with no reasoning involved, but they are a critical step towards AGI nonetheless.

It is in this second step where the forefront of AI research is right now. The first approaches building episodic reactive behaviours include recurrent neural networks~\cite{Pineda87} and long short-term memories~\cite{hs97}. The most recent development is the Transformer architecture~\cite{vspujgkp17}, which has become the base architecture of GPTs and foundation models. These approaches have managed to achieve groundbreaking milestones, such as breaking the DNA code~\cite{npftwbmprberb23,dgmcgodtas24,zjlddl24} or passing the Turing Test~\cite{Sejnowski23}. However, they still lack the development of reasoning mechanisms~\cite{ckp23,hcmzysz24}, although some reasoning-like behaviour has been recently achieved using LLMs and textual chain of thoughts~\cite{ssfw+23,avllzy24,ygkgr24,rphb24}.

Recently, a novel approach developing cognitive architectures from mere inputs has been proposed, called Synthetic Cognition~\cite{irga24}. However, in the path to develop these cognitive architectures, to date, the proposal has been developed only to produce instantaneous reactive behaviours~\cite{iarga24}. In this study, we aim to explore how Synthetic Cognition can be extended to develop episodic reactive behaviours, to implement the Declarative Metacluster presented in~\cite{irga24}.

We decided to start with the simplest approach to deal with episodes or sequences: treating the sequence as a window in which each element of the sequence corresponds to a different timestamp. In other words, considering the input to be composed of the instantaneous element of the present time plus the instantaneous elements of the previous \emph{n} times, in what we can call a \emph{context-aware input}. This is in fact the Transformers' approach: they receive a window corresponding to the current element of the sequence and the previous \emph{n} elements. Those \emph{n} previous elements are called \emph{the context window}, and they allow the Transformer to provide a context-aware answer.

To replicate Transformer's success in processing sequences, we took the first implementation of Synthetic Cognition (Unsupervised Cognition~\cite{iarga24}), which only deals with instantaneous inputs, and provided it with context-aware inputs. Thus, the algorithm is exactly the same that deals with instantaneous inputs, but this time dealing with sequences because the inputs are provided with their corresponding context windows. The goal of this test is twofold: on one hand, it will allow us to develop context-aware methods using Synthetic Cognition's approach, and on the other hand, it will evaluate its robustness as a primitive-based framework to build cognitive architectures. If we are able to deal with sequences just tweaking the inputs the system receives, then we can integrate such input changes into the whole system.

Given the inspiration in Transformers, and their current status as state-of-the-art, we decided to test our approach in a benchmark against Transformer-based models. Specifically, we used a recently published benchmark~\cite{fwzhzwlww24} that compares three DNA foundation models over a set of 44 DNA sequence classification datasets. This task is relevant because decoding DNA sequences to understand epigenetic patterns, transcriptional regulation, and/or disease associations provides useful insights to doctors when treating or preventing illnesses.

In our experiments, with a small context window, we managed to outperform DNA foundation models in more datasets than each one of them, thus obtaining the highest mark in more datasets. Moreover, these results were obtained without pre-training, unlike the foundation models that needed huge pre-training before being fine-tuned to solve each of the datasets of the benchmark.

These results show the potential of Synthetic Cognition to beat not only traditional Machine Learning methods in an unsupervised learning setting~\cite{iarga24}, but also more advanced methods, such as Transformers, in an episodic setting. This is a fundamental stone in the path towards AGI, as episodic reactive behaviour is a fundamental building block over which to develop any kind of reasoning. The subsequent steps will include the development of reasoning mechanisms over the learned episodes; however, this is a matter of future work.

The remainder of this paper is organized as follows. Section~\ref{sec:relwork} introduces previous work related to our research. Section~\ref{sec:epi} presents our episodic setup for Synthetic Cognition. Section~\ref{sec:exp} details the experiments that were performed. Section~\ref{sec:dis} explores the implications of this study. Section~\ref{sec:lim} outlines the limitations of our proposed method. Finally, Section~\ref{sec:conc} highlights the conclusions of the study.

\section{Related Work}\label{sec:relwork}
In the field of Artificial Intelligence, the current state-of-the-art method for dealing with sequences is the Transformer architecture. This architecture, based on the widely popular artificial neural networks, combines a set of neurons focused on identifying the input with a set of neurons focused on setting attention along the input. As it is based on neural networks, it is a weight-based algorithm and thus requires enormous amounts of data to be properly tuned for the task at hand. Given this data constraint, a huge field has been developed to build what has been called \emph{foundation models}. These models are Transformer architectures trained with huge datasets to properly tune the network weights to a given knowledge domain. Subsequently, to solve a specific task, additional layers of neurons are added. These layers take the output of the foundational model as input and are fine-tuned for the specific task at hand. The idea of this setup is that the foundation model has learned to identify elements of the knowledge domain and that the last layers, fine-tuned to the new task, will work better owing to the transformation produced by the foundation model.

With the advent of the new millennium, advances in biotechnology have facilitated a precipitous drop in DNA sequencing costs. Because of this, a flood of genetic data has emerged ready to be capitalized on by translational scientists, from clinical applications on humans to biotechnological developments of commercial crops. However, decoding DNA information to understand epigenetic patterns, transcriptional regulation, and disease associations remains the main bottleneck for leveraging potential applications. 
Recently, DNA foundation models that use the transformer's technology have emerged: DNABERT-2~\cite{zjlddl24}, HyenaDNA~\cite{npftwbmprberb23} and Nucleotide Transformer (v2)~\cite{dgmcgodtas24}. These models are pre-trained on massive genomic datasets, such as the Human Genome~\cite{sghbc+17} for all models, whereas Nucleotide Transformer (v2) and DNABERT-2 have received additional training with the output of the 1000 Genomes Project~\cite{bezb+22} and $135$ non-human species, respectively. All these datasets are large enough to build foundational models, and therefore, are orders of magnitude larger than the datasets from the benchmark presented in this paper.

\section{Episodic Cognition}\label{sec:epi}
Inspired by the Synthetic Cognition framework presented in ~\cite{irga24}, Unsupervised Cognition was developed~\cite{iarga24}. This was an initial implementation of Synthetic Cognition that addressed the unsupervised learning problem, and it was successfully compared with other unsupervised learning algorithms. In this regard, it only implemented the so-called Motoperceptive Metacluster~\cite{irga24}. This Metacluster builds a tree-like knowledge representation composed of abstract representations of the learned inputs.

It builds representations by aggregating similar inputs, with the goal of modelling the underlying domain, and then provides the most similar representation when asked to identify a new input. These abstractions are organised in a tree-like hierarchical structure where higher order representations are compositions of their children.

In the original Unsupervised Cognition, each input is composed of different features. Our proposal is to use the same algorithm, but with each input composed of different timestamp. That is, our proposal to deal with sequences is that each input will be composed of multiple timestamps (e.g., sequence elements), and the rest of the algorithm remains the same. We encourage reading~\cite{iarga24} to fully understand how Unsupervised Cognition works.

To build such inputs, we apply a window to the sequence with a stride defining the number of elements the window moves to produce the following input. We set the stride to $1$ by default. In other words, the first input is the set of $n$ consecutive elements of the sequence starting with the first element of the sequence, and the second input is the set of $n$ consecutive elements of the sequence starting with the second element of the sequence. And so on.

\section{Experiments}\label{sec:exp}
In this section, we present the experiment that we performed against Transformer models to evaluate the suitability of our proposal for dealing with sequences. We used a benchmark for DNA Sequence Classification~\cite{fwzhzwlww24} and evaluated our results against those produced by three DNA sequence foundation models: DNABERT-2~\cite{zjlddl24}, HyenaDNA~\cite{npftwbmprberb23} and Nucleotide Transformer (v2)~\cite{dgmcgodtas24}. To ensure that we took a benchmark in which processing inputs as sequences was crucial, we tested such a benchmark with Unsupervised Cognition~\cite{iarga24} (Synthetic Cognition's instantaneous version) and verified that the obtained results were disastrous. Thus, it is clear that we need an improved version of Synthetic Cognition.

\subsection{The Benchmark}
Synthetic Cognition was evaluated against a comprehensive benchmark introduced by the University of Texas MD Anderson Cancer Center~\cite{fwzhzwlww24}, comprising $57$ DNA sequence classification datasets spanning a wide range of biological contexts and species. These datasets cover tasks such as finding DNA sequences prone to undergo epigenetic modifications (e.g., 4mC, 5mC, and 6mA), the identification of DNase-I hypersensitive sites, and other regulatory related regions, such as promoters, enhancers, and splice sites across different organisms. Despite this diversity, the core challenge across all datasets is the same: predicting a biological trait or origin from raw DNA sequences alone, while assessing both intra-species and across-species generalization capabilities. The only exception is the classification of COVID-19 viral strains based on genomic fragments. 

To ensure that the evaluation remains fair and realistic, the benchmark employs both curated datasets used in the original evaluation of foundation models~\cite{zjlddl24,npftwbmprberb23,dgmcgodtas24} and newly gathered public datasets to verify the quality and minimize redundancy (e.g., in epigenetic trait detection tasks, sequences with high similarity were removed to reduce bias).

Related to the type of classification task, the sequences vary considerably in length in terms of base pairs (bp). Some datasets have uniform sequence lengths, such as the $41$bp inputs used in the 4mC/5mC/6mA detection. Others exhibit substantial variations, including promoter datasets from human cell lines, which can span up to $3000$bp. This diversity in terms of DNA sequence length allows us to test for possible effects of input size on performance. The whole size of all datasets is displayed in Table~\ref{tab:data}.

It should be pointed out that, among the $57$ datasets, $15$ were grouped for evaluation purposes, specifically, the five mouse functional motif datasets and ten yeast epigenetic mark datasets. For these grouped tasks, an average score is computed across the datasets in the group in order to provide a single aggregated metric. This largely reduces the number of individual evaluation scores from $57$ to $44$, thus simplifying performance comparisons while preserving task diversity.

\begin{table}[!t]\centering
\caption{Benchmark Datasets (ordered by total train size)}\label{tab:data}
\scriptsize
\begin{tabular}{| l | r | r | r | r | r |}
\hline
\textbf{Dataset} &\textbf{Tr. Smpl.} &\textbf{Tst. Smpl.} &\textbf{Max. Len.} &\textbf{Avg. Len.} &\textbf{Total Tr. Size} \\
\hline
\hline
Promoter B\_amyloliquefaciens &1,483 &636 &40 &40 &59,320 \\
\hline
5-methylcytosin(5mC) &2,344 &2,344 &41 &41 &96,104 \\
\hline
DNase\_I Hypersensitive &711 &306 &275 &243 &172,773 \\
\hline
Mouse TFBS 4 &1,904 &239 &101 &101 &192,304 \\
\hline
Mouse TFBS 3 &2,620 &328 &101 &101 &264,620 \\
\hline
Promoter R\_capsulatus &7,406 &3,175 &40 &40 &296,240 \\
\hline
Promoter TATA 70 bps &4,904 &613 &70 &70 &343,280 \\
\hline
E.Coli 4mC &8,681 &3,721 &41 &41 &355,921 \\
\hline
Mouse TFBS 1 &6,478 &810 &101 &101 &654,278 \\
\hline
N6-methyladenosine(6mA) &18,336 &18,334 &41 &41 &751,776 \\
\hline
Promoter Arabidopsis TATA &3,063 &1,313 &251 &251 &768,813 \\
\hline
G.Pickeringii 4mC &24,053 &10,309 &41 &41 &986,173 \\
\hline
Promoter TATA 300 bps &4,904 &613 &300 &300 &1,471,200 \\
\hline
Mouse TFBS 5 &15,064 &1,883 &101 &101 &1,521,464 \\
\hline
TFBS Data 3 &19,000 &1,000 &101 &101 &1,919,000 \\
\hline
TFBS Data 5 &19,000 &1,000 &101 &101 &1,919,000 \\
\hline
Promoter Arabidopsis NonTATA &8,267 &3,543 &251 &251 &2,075,017 \\
\hline
G.Subterraneus 4mC &63,567 &27,243 &41 &41 &2,606,247 \\
\hline
TFBS Data 4 &27,294 &1,000 &101 &101 &2,756,694 \\
\hline
Promoter NonTATA 70 bps &42,452 &5,307 &70 &70 &2,971,640 \\
\hline
Enhancer &14,968 &400 &200 &199 &2,978,632 \\
\hline
Enhancer Strength &14,968 &400 &200 &199 &2,978,632 \\
\hline
TFBS Data 2 &30,672 &1,000 &101 &101 &3,097,872 \\
\hline
Promoter NHEK &8,170 &2,044 &2,400 &400 &3,268,000 \\
\hline
TFBS Data 1 &32,378 &1,000 &101 &101 &3,270,178 \\
\hline
Promoter All 70 bps &47,356 &5,920 &70 &70 &3,314,920 \\
\hline
C.Elegans 4mC &84,926 &36,398 &41 &41 &3,481,966 \\
\hline
D.Melanogaster 4mC &126,466 &54,200 &41 &41 &5,185,106 \\
\hline
Mouse TFBS 2 &53,952 &6,745 &101 &101 &5,449,152 \\
\hline
Yeast Epigenetic Marks 9 &11,679 &1,461 &500 &500 &5,839,500 \\
\hline
Yeast Epigenetic Marks 1 &11,971 &1,497 &500 &500 &5,985,500 \\
\hline
A.Thaliana 4mC &156,697 &67,157 &41 &41 &6,424,577 \\
\hline
Promoter NonTATA 251 bps &27,097 &9,034 &251 &251 &6,801,347 \\
\hline
Enhancer Cohn &20,843 &6,948 &500 &500 &10,421,500 \\
\hline
Splice Site Type NT &27,000 &3,000 &400 &400 &10,800,000 \\
\hline
Yeast Epigenetic Marks 8 &22,224 &2,779 &500 &500 &11,112,000 \\
\hline
Yeast Epigenetic Marks 7 &23,069 &2,884 &500 &500 &11,534,500 \\
\hline
Donor &19,775 &2,198 &600 &600 &11,865,000 \\
\hline
Acceptor &19,961 &2,218 &600 &600 &11,976,600 \\
\hline
Yeast Epigenetic Marks 5 &24,545 &3,069 &500 &500 &12,272,500 \\
\hline
Yeast Epigenetic Marks 4 &25,341 &3,168 &500 &500 &12,670,500 \\
\hline
Promoter NonTATA 300 bps &42,452 &5,307 &300 &300 &12,735,600 \\
\hline
Promoter Hela-S3 &11,736 &2,936 &2,999 &1,113 &13,062,168 \\
\hline
Yeast Epigenetic Marks 2 &26,438 &3,305 &500 &500 &13,219,000 \\
\hline
Yeast Epigenetic Marks 10 &27,275 &3,410 &500 &500 &13,637,500 \\
\hline
Yeast Epigenetic Marks 3 &27,904 &3,488 &500 &500 &13,952,000 \\
\hline
Promoter All 300 bps &47,356 &5,920 &300 &300 &14,206,800 \\
\hline
Splice Site Type DNABERT-2 &36,496 &4,562 &400 &400 &14,598,400 \\
\hline
Yeast Epigenetic Marks 6 &29,439 &3,680 &500 &500 &14,719,500 \\
\hline
Coding &75,000 &25,000 &200 &200 &15,000,000 \\
\hline
Human vs worm &75,000 &25,000 &200 &200 &15,000,000 \\
\hline
Promoter HUVEC &11,928 &2,982 &2,997 &1,267 &15,112,776 \\
\hline
Promoter GM12878 &10,992 &2,750 &2,999 &1,622 &17,829,024 \\
\hline
Enhancer Ensembl &123,872 &30,970 &573 &269 &33,321,568 \\
\hline
Open chromatin region &139,804 &34,952 &593 &315 &44,038,260 \\
\hline
Regulatory Region Type &150,000 &57,713 &802 &401 &60,150,000 \\
\hline
Covid Variants &73,335 &9,168 &999 &999 &73,261,665 \\
\hline
\end{tabular}
\end{table}

\subsection{The Experimental Setup}
In the benchmark experiments, for each dataset, the authors took each of the DNA foundation models, processed the sequences with them to obtain zero-shot sequence embeddings, and then trained a random forest using 5-fold cross-validation. Then, the trained random forest was used to perform the final classification over the test set, and the Area Under the Curve (AUC) was computed.

In our case, because our proposal does not require additional methods to perform classification, we have a simpler pipeline. For each dataset, we only used the training set and trained our algorithm with it. Then, we evaluated the test set with the resulting model to produce the classification labels, and computed the AUC over them. As our algorithm does not have hyper-parameters, we do not need 5-fold cross-validation either, and we evaluate directly over the test set.

The only quirk of our proposal is that we process multiple inputs for each sequence (one per window), and thus obtain multiple outputs. To harmonize all those outputs, we decided to select the most repeated class as the final answer, computing the probability of each possible class based on their frequency in the set of outputs. This could have slightly hampered our results due to the fact that not all inputs would have enough discriminatory information to properly classify the whole sample, but it is a trade-off that we had to make.

Finally, our proposal was tested with a window of $n=5$ elements due to resource limitations, but with larger windows, we know we obtain better results.
For comparison, the other methods used windows on the order of thousands of elements. We performed our experiments with our proposal on an Ubuntu laptop with an Intel Core i9-13900HX at 2.60GHz with 32 cores, 32Gb of memory, and an NVIDIA GeForce RTX 4060 with 8Gb of VRAM. The results of the other methods were obtained using the aforementioned benchmark.

\subsection{The Results}
After executing the experiments, we obtained the results listed in Table~\ref{tab:res}. There, we can see how, although our proposal is not better for all datasets, it is better in $36.36\%$ of them (with an average rank of $2.295$), with DNABERT-2 being better in $36.36\%$ (with an average rank of $1.977$), Nucleotide Transformer (v2) being better in $22.73\%$ (with an average rank of $2.477$), and HyenaDNA being better in a merely $4.55\%$ of datasets (with an average rank of $3.159$).

\begin{table}[!t]\centering
\caption{Benchmark Results (ordered by total train size)}\label{tab:res}
\scriptsize
\begin{tabular}{| l | l | l | l | l |}
\hline
\textbf{Dataset} &\textbf{DNABERT-2} &\textbf{Nucl. Trans.} &\textbf{HyenaDNA} &\textbf{Synth. Cogn.} \\
\hline
\hline
Promoter B\_amyloliquefaciens &0.856 &0.797 &0.688 &\bf 0.882 \\
\hline
5-methylcytosin(5mC) &0.678 &\bf 0.713 &0.604 & 0.674 \\
\hline
DNase\_I Hypersensitive &0.815 &0.806 &0.787 &\bf 0.835 \\
\hline
Promoter R\_capsulatus &0.661 &0.668 &0.602 &\bf 0.709 \\
\hline
Promoter TATA 70 bps &0.809 &\bf 0.872 &0.702 &0.785 \\
\hline
E.Coli 4mC &0.567 &\bf 0.579 &0.579 & 0.5 \\
\hline
N6-methyladenosine(6mA) &0.731 &0.752 &0.681 &\bf 0.758 \\
\hline
Promoter Arabidopsis TATA &0.903 &0.855 &0.82 &\bf 0.94 \\
\hline
G.Pickeringii 4mC &0.587 &\bf 0.607 &0.603 &0.5 \\
\hline
Promoter TATA 300 bps &0.698 &0.694 &\bf 0.717 &0.629 \\
\hline
TFBS Data 3 &0.744 &0.715 &0.715 &\bf 0.808 \\
\hline
TFBS Data 5 &0.681 &0.647 &0.636 &\bf 0.865 \\
\hline
Promoter Arabidopsis NonTATA &0.891 &0.85 &0.814 &\bf 0.94 \\
\hline
G.Subterraneus 4mC &\bf 0.588 &0.581 &0.577 &0.5 \\
\hline
TFBS Data 4 &0.732 &\bf 0.764 &0.732 &0.733 \\
\hline
Promoter NonTATA 70 bps &0.816 &\bf 0.835 &0.803 &0.825 \\
\hline
Enhancer &0.863 &\bf 0.879 &0.833 &0.801 \\
\hline
Enhancer Strength &0.515 &0.471 &0.485 &\bf 0.724 \\
\hline
TFBS Data 2 &0.834 &0.836 &0.842 &\bf 0.892 \\
\hline
Promoter NHEK &\bf 0.912 &0.855 &0.854 &0.886 \\
\hline
TFBS Data 1 &0.817 &0.824 &0.83 &\bf 0.86 \\
\hline
Promoter All 70 bps &0.803 &\bf 0.822 &0.769 &0.801 \\
\hline
C.Elegans 4mC &0.587 &0.594 &0.583 &\bf 0.626 \\
\hline
D.Melanogaster 4mC &0.604 &0.611 &0.57 &\bf 0.639 \\
\hline
A.Thaliana 4mC &0.59 &0.6 &0.557 &\bf 0.604 \\
\hline
Promoter NonTATA 251 bps &\bf 0.861 &0.834 &0.853 &0.821 \\
\hline
Mouse TFBS (all) &0.7 &0.722 &0.624 &\bf 0.825 \\
\hline
Enhancer Cohn &\bf 0.792 &0.728 &0.733 &0.746 \\
\hline
Splice Site Type NT &0.712 &\bf 0.725 &0.71 &0.574 \\
\hline
Donor &\bf 0.823 &0.636 &0.626 &0.651 \\
\hline
Acceptor &\bf 0.793 &0.632 &0.67 &0.616 \\
\hline
Promoter NonTATA 300 bps &\bf 0.938 &0.91 &0.818 &0.839 \\
\hline
Promoter Hela-S3 &\bf 0.971 &0.909 &0.9 &0.937 \\
\hline
Promoter All 300 bps &\bf 0.897 &0.855 &0.797 &0.814 \\
\hline
Splice Site Type DNABERT-2 &\bf 0.608 &0.607 &0.607 &0.5 \\
\hline
Coding &\bf 0.915 &0.863 &0.885 &0.874 \\
\hline
Human vs worm &\bf 0.946 &0.919 &0.837 &0.921 \\
\hline
Promoter HUVEC &\bf 0.974 &0.912 &0.906 &0.939 \\
\hline
Promoter GM12878 &\bf 0.964 &0.878 &0.884 &0.925 \\
\hline
Enhancer Ensembl &0.947 &\bf 0.95 &0.944 &0.704 \\
\hline
Open chromatin region &\bf 0.685 &0.657 &0.665 &0.638 \\
\hline
Regulatory Region Type &0.63 &0.555 &\bf 0.702 &0.621 \\
\hline
Covid Variants &0.446 &0.43 & 0.449 &\bf 0.56 \\
\hline
Yeast Epigenetic Marks (all) &\bf 0.734 &0.643 &0.665 &0.704 \\
\hline
\end{tabular}
\end{table}

A remarkable result from this test is that our proposal obtains better results for all tasks related to the detection of epigenetic motifs. In fact, the only tasks in which we sometimes get worse results are those concerning the detection of functional motifs. Moreover, in the only task regarding the identification of COVID-19 strains based on viral genome fragments, our proposal obtained much better results than the alternatives. However, as we are not experts on DNA Sequences, we are not able to provide more insights into why these differences between tasks.

Finally, we would like to signal that our results are not associated with better performance on smaller datasets. Although it is true that we beat the alternatives in the smaller datasets, we consider this to be a consequence of the smaller window size. In fact, for one of the largest datasets (the COVID-19 dataset), we also obtained better scores than the alternatives. It is true that the bigger the dataset, the bigger the window size; however, adjusting the window size is sufficient for our proposal to beat the alternatives.

\begin{figure}
    \centering
    \includegraphics[width=0.75\linewidth]{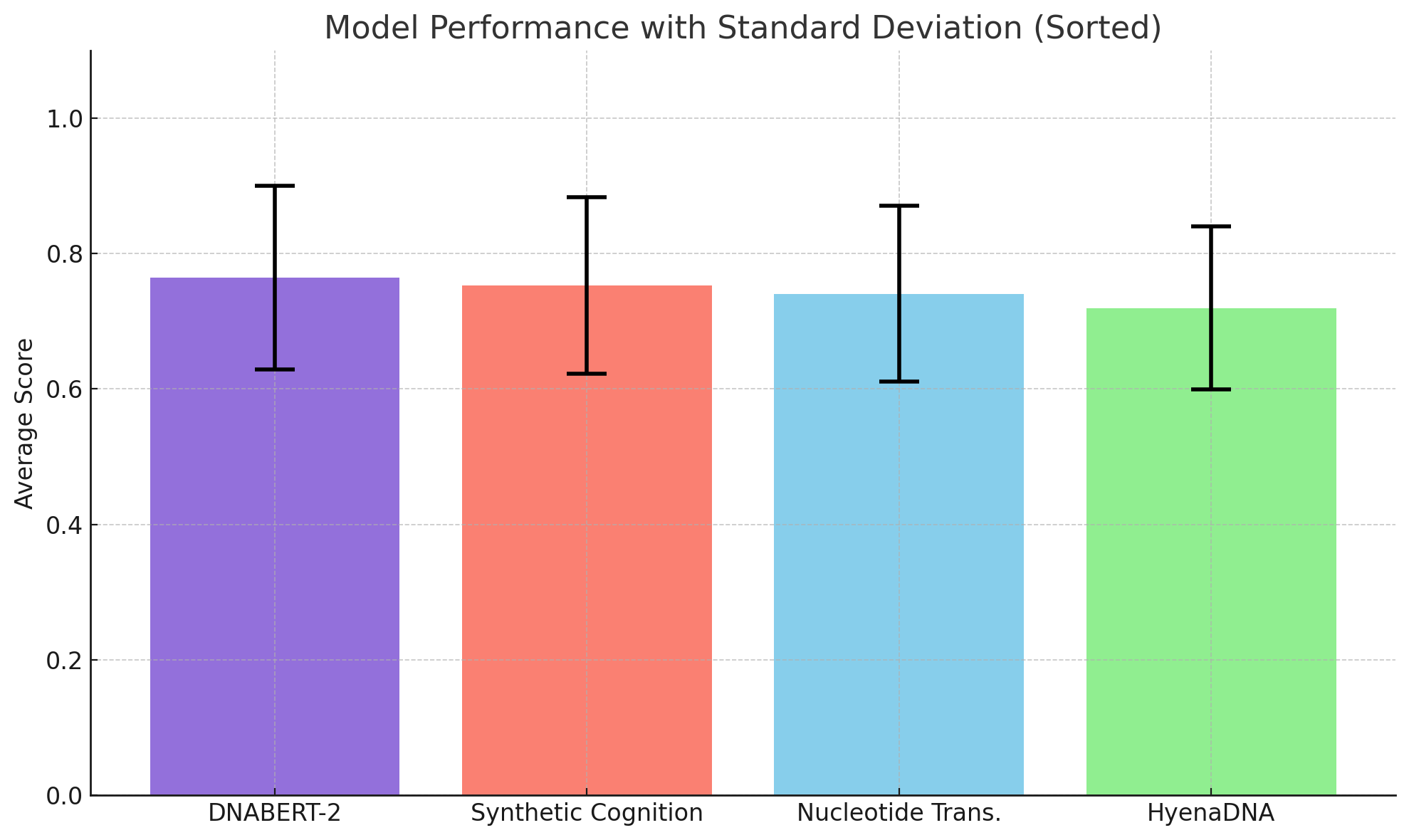}
    \caption{Average score per model with standard deviation as error bars.}
    \label{fig:avgandstd}
\end{figure}

In fact, doing a brief exploration of bigger windows, we were able to beat the alternatives also for the ``E.Coli 4mC'' (window size $= 10$, score $= 0.605$) and ``5-methylcytosin(5mC)'' (window size $= 11$, score $= 0.75$) datasets. This updates the results as follows: our proposal is better in $40.91\%$ of the datasets with an average rank of $2.182$, DNABERT-2 is better in $36.36\%$ with an average rank of $2.023$, Nucleotide Transformer (v2) is better in $18.18\%$ with an average rank of $2.532$, and HyenaDNA is better in only $4.55\%$ of the datasets with an average rank of $3.182$. The average score and its standard deviation is displayed in Figure~\ref{fig:avgandstd}. With these results, it is clear that our proposal is better suited to deal with DNA Sequence Classification tasks.

\section{Discussion}\label{sec:dis}
In this Section, we discuss two matters: why we are not winning in all datasets and what are the effects of pre-training in our model.

Regarding the fact that we do not obtain better scores than the alternatives in all datasets, we would like to point out that these datasets encompass very different tasks, each one with its own quirks and idiosyncrasy. However, due to our limited resources, we solved the datasets in bulk. That is, the configuration for all datasets was the same and we used a small window size. We do not consider this to be a problem because our achievements are already good proof that our proposal is a better alternative to Transformer DNA foundation models. However, multiple actions were available to improve the results. For instance, we could further extend the time context of our algorithm to increase the context-awareness of our answers. This is critical, particularly for datasets with very long sequences. Another alternative would be to change the method by which we harmonize the multiple outputs of our algorithm, that is, in some cases, it could be better to decide that one class is the default, and the other class is selected as soon as one output says it has recognized that other class.

Regarding the fact that we do not have pre-training, we would like to point out that this is an advantage of our proposal. Transformer DNA foundation models require large amounts of data for training, as explained in Section~\ref{sec:relwork}. This, in turn, makes these models time-intensive and resource-hungry. In contrast, our proposal only needs a fine-tuning dataset, requiring hundreds of thousands of less resources and time. Moreover, our proposal is better suited for the task at hand because it has only information about such a task. In fact, performing a huge pre-training for our model has the potential to be counterproductive, as more information can lead to more ambiguity and the associated worsening of results.

Pre-training makes sense for a Transformer architecture because there are many weights that have to be properly tuned, and thus, a huge amount of data is necessary. However, in our case, as we do not have weights to tune, but instead we build representations, any unrelated information we process is useless, as it will never be used when performing the task at hand. Moreover, any closely related but ambiguous information has the potential of confusing the model.

\section{Limitations}\label{sec:lim}
Regarding the limitations of our proposal, we mainly have one: memory consumption. Our algorithm builds a representation of each input it processes during training. Thus, each training input consumes memory. In addition, more representations are generated to effectively construct the abstractions of the inputs, which is crucial for our algorithm to handle new, unseen samples. However, this approach comes with the trade-off of increased memory consumption. Thus, our proposed method has a significant memory consumption problem. As we represent inputs as SDRs, although big, these memory requirements still allow us to process hundreds of thousands of samples; however, they impose a limitation on the size of our models. We are working on mechanisms to address this problem, from pruning unused or redundant representations to optimize memory use, but they are a matter of future work.

This limitation has a critical consequence: we cannot deal with Natural Language Processing (NLP) tasks, at least for now. For this reason, the target dataset for our experiments was DNA Sequencing because the realm of words is relatively small, thus building a limited number of representations. However, in the NLP realm, the number of words is massive, and most of them are associated with other words (i.e., synonyms), which results in our algorithm building enormous numbers of representations that limit our capability of processing such tasks. However, we are working to address these problems, and we expect those efforts to allow us to deal with this kind of task, which is more well regarded in the world of sequence processing.

\section{Conclusions}\label{sec:conc}
Dealing with episodes is a fundamental task for any method that aims to develop Artificial General Intelligence. To date, Transformer architecture is the best approach for dealing with episodes. However, they still have some limitations in developing their reasoning skills. Recently, a new approach for building cognitive architectures based on literal inputs has been proposed. However, such an approach has only been developed to deal with instantaneous reactive behaviour. In this paper, we have proposed a mechanism for such approach to deal with episodic reactive behaviours, that is, with sequences.

We have tested our approach over a DNA sequence classification benchmark, to compare our proposal with the Transformer architecture. In fact, we compare against three widely known foundation models designed to learn representations from DNA sequences that encode their biological functions. In our experiments, we proved that our proposed method is better suited for dealing with DNA sequence classifications, showing that we obtained the best score for more datasets than any other method. Moreover, we managed to obtain such results without the costly pre-training that Transformer foundation models require.

In future work, we would like to test our approach on more benchmarks, such as the one that came with Nucleotide Transformer (v2)~\cite{dgmcgodtas24}. We would also like to integrate our approach into a whole Synthetic Cognition system, creating a two-tier model with Semantic Memory (Motoperceptive Metacluster) and Declarative Memory (Declarative Metacluster). In our proposal, we would like to explore ways to reduce memory consumption. Finally, we would like to test our proposal over Natural Language Processing tasks to further compare it with Transformers, maybe allowing to build Large Language Models with it.

\subsubsection{Acknowledgments.}
We would like to thank Daniel Pinyol and Pere Mayol for their insightful discussions on the topic. This work has been supported by the Torres-Quevedo grant PTQ2023-012986 funded by the MCIU/AEI /10.13039/501100011033.

\subsubsection{Disclosure of Interests.}
The authors have no competing interests to declare relevant to the content of this article.

\bibliographystyle{splncs04}
\bibliography{biblio}

\end{document}